\documentclass[
]{ceurart}

\usepackage{listings}

\usepackage{caption}
\usepackage{subcaption}
\begin{document}

\copyrightyear{2022}
\copyrightclause{Copyright for this paper by its authors.
  Use permitted under Creative Commons License Attribution 4.0
  International (CC BY 4.0).}

\conference{AIC 2023 (9th International Workshop on Artificial Intelligence and Cognition)}

\title{Exploring Temporal Representation in Neural Processes for Multimodal Action Prediction}

\author[1,2]{Marco Gabriele Fedozzi}[%
orcid=0000-0001-7140-0106,
email=marco.fedozzi@iit.it,
]
\cormark[1]
\address[1]{DIBRIS Department, University of Genoa, Via All'Opera Pia 13 16145 Genoa, Italy}
\address[2]{CONTACT Unit, Italian Institute of Technology, Via Enrico Melen 83, 16152 Genoa, Italy}

\author[3]{Yukie Nagai}[%
orcid=0000-0003-4794-0940,
email=nagai.yukie@mail.u-tokyo.ac.jp,
]
\address[3]{International Research Center for Neurointelligence, The University of Tokyo 7-3-1 Hongo, Bunkyo-ku, Tokyo 113-0033, Japan}

\author[1,2]{Francesco Rea}[%
orcid=0000-0001-8535-223X,
email=francesco.rea@iit.it,
url=http://www.francescorea.eu/curriculum.html,
]
\author[1,2]{Alessandra Sciutti}[%
orcid=0000-0002-1056-3398,
email=alessandra.sciutti@iit.it,
]

\cortext[1]{Corresponding author.}

\begin{abstract}
 Inspired by the human ability to understand and predict others, we study the applicability of Conditional Neural Processes (CNP) to the task of self-supervised multimodal action prediction in robotics. Following recent results regarding the ontogeny of the Mirror Neuron System (MNS), we focus on the preliminary objective of self-actions prediction. We find a good MNS-inspired model in the existing Deep Modality Blending Network (DMBN), able to reconstruct the visuo-motor sensory signal during a partially observed action sequence by leveraging the probabilistic generation of CNP. After a qualitative and quantitative evaluation, we highlight its difficulties in generalizing to unseen action sequences, and identify the cause in its inner representation of time. Therefore, we propose a revised version, termed DMBN-Positional Time Encoding (DMBN-PTE), that facilitates learning a more robust representation of temporal information, and provide preliminary results of its effectiveness in expanding the applicability of the architecture. DMBN-PTE figures as a first step in the development of robotic systems that autonomously learn to forecast actions on longer time scales refining their predictions with incoming observations.
\end{abstract}


\maketitle

\section{Introduction}

Understanding the motions and intentions of others has garnered significant research interest due to its relevance to human cognition. The Mirror Neuron System (MNS), discovered in monkeys \cite{gallese1996action} and its functional analogue in humans \cite{mukamel2010single}, provides insights into the neural mechanisms underlying social cognitive abilities \cite{oztop2013mirror}.
The MNS plays a crucial role in the Simulation Theory (ST) of mindreading, suggesting that mentalization arises from an internal simulation of observed agents' actions \cite{gallese1998mirror}, after their  conversion from an allocentric to the egocentric point of view. Although different computational models have been proposed to explain this ability, few present high biological plausibility \cite{schrodt2015embodied}. Recent results demonstrate that the MNS responds to self-actions too, coordinating the forward and inverse models used for action planning and learning \cite{bonini2017extended}. This property is assumed to play a central role during the development of the mirroring abilities in infants \cite{giudice2009programmed, gerson2014learning}, linking outcomes to self-motion before focusing on other agents, allowing to learn a model of both the self and the outer world.
% the ability to prdict the outcomes of ones' own actions is considered at the cornerstone of our ability to form world models to interacti with the outside envirnoment.
The predictive abilities of the MNS on self- and others' actions highlight the human brain's predictive nature, aligning with the Predictive Coding (PC) theory \cite{rao1999predictive, spratling2017review, millidge2021predictive}.
Following the claim that human-like cognition in robots would ease their integration and transparency in our society  \cite{sandini2018social, cangelosi2022cognitive}, MNS-inspired robotic models were designed to understand and predict the outer world. 
%These models have shifted from symbolic \cite{demiris2006hierarchical} to emergent approaches, better aligned with developmental studies \cite{copete2016motor} and more apt at representing multimodal signals \cite{droniou2015deep}.
In the following sections, we will explore existing architectures, before focusing on the most promising one \cite{seker2022imitation} in Section \ref{sec:related-work}, and proposing enhancements to mitigate its shortcomings in Section \ref{sec:evaluation}.
% - prediction of visual information is strictly correlated with the Video Prediction task
%   in CV, although they lack the proprioceptive information which could ease learning (cite MNs
%   studies on that, as proprioception during development gives a robust signal on which to ground 
%   self-actions)

\section{Related Work} \label{sec:related-work}

A MNS-inspired robotic model, to be effective, should: operate on multimodal signals, connecting observations with motor plans for enaction and understanding \cite{meo2021multimodal}; predict the outcomes of self-actions, to acquire a model of the outside world \cite{taniguchi2023world}; connect third-person to first-person perspective, going beyond geometric mapping and accounting for individual differences \cite{hunnius2014you}.
Existing solutions fall short in simultaneously satisfying all those desired properties.
%Learning the correlation between actions and observations in a manner similar to the MNS endows robots with two favorable properties. It enables the formulation of simple plans by generating motor sequences from the observation of a goal posture \cite{meo2021multimodal, seker2019conditional, seker2022imitation}, an more robust approach than classical planners to noise and unknown properties. Additionally, extending this forecasting ability to other agents allows the robot to predict their goals and learn new imitative behaviors \cite{copete2016motor, zambelli2020multimodal}. 

The architectures in \cite{zambelli2020multimodal, seker2019conditional} operate on highly-preprocessed data, limiting the ability to learn deep correlations between modalities.
Existing predictive models \cite{copete2016motor, meo2021multimodal, zambelli2020multimodal} are autoregressive, requiring multiple steps to generate predictions further in time, thus suffering from the accumulation of errors. 
The Deep Modality Blending Network (DMBN) \cite{seker2022imitation} overcomes these limitations by leveraging Conditional Neural Processes (CNP) \cite{garnelo2018conditional} in order to predict in parallel points at any distance in the future from the current observations.
Lastly, existing architectures do not address the self-other alignment problem, operating from fixed third-person \cite{meo2021multimodal} or first-person perspectives \cite{copete2016motor, zambelli2020multimodal}. Interestingly, \cite{seker2022imitation} suggests that the DMBN might carry out such perspective shift without specific training, though further investigation is needed.

For those reasons, the DMBN \cite{seker2022imitation} is chosen here as a promising architecture for modeling MNS-like properties (refer to the original work \cite{seker2022imitation} for a more in-depth explanation).

\subsection{Conditional Neural Processes} \label{subsec:CNP}
% brief explanation on how they work

Neural Processes (NP) \cite{garnelo2018neural, dubois2020npf} are a family of Artificial Neural Networks (ANNs) combining traditional ANNs with Gaussian Processes (GP) \cite{seeger2004gaussian}. They operate on sets of data and learn to model the distribution of a target set conditioned on a context set. The context points are encoded individually and combined using a set operator (e.g. averaging) to generate a \textit{representative element} for the set. Positional information is added to this element before decoding, in order to reconstruct different targets.
Conditional Neural Processes (CNPs) \cite{garnelo2018conditional} are a branch of NP that factorize the dependency between the target and context sets, sacrificing complexity to ease the training, while still representing the reconstruction uncertainty on the target set.

% add pictures of both CNPs and LNPs

\subsection{Deep Modality Blending Networks}

\begin{figure*}[t]
  \centering
      \begin{subfigure}[b]{0.45\linewidth}
        \centering
          \includegraphics[width=\linewidth]{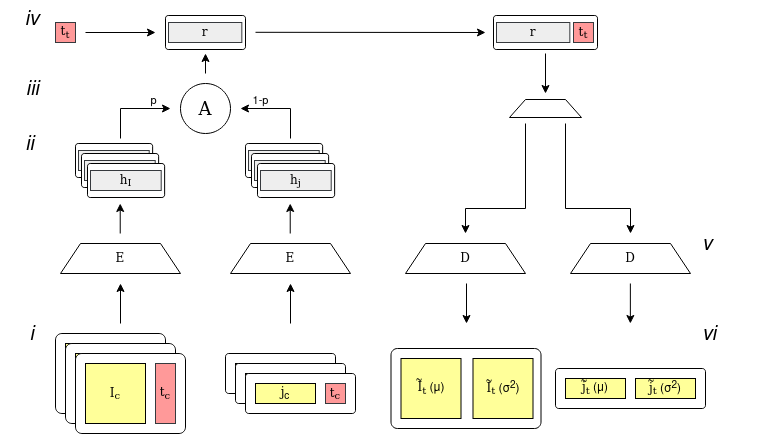}
          \subcaption{Original DMBN architecture}
          \label{fig:dmbn-architecture}
      \end{subfigure}
      \begin{subfigure}[b]{0.45\linewidth}
        \centering
          \includegraphics[width=\linewidth]{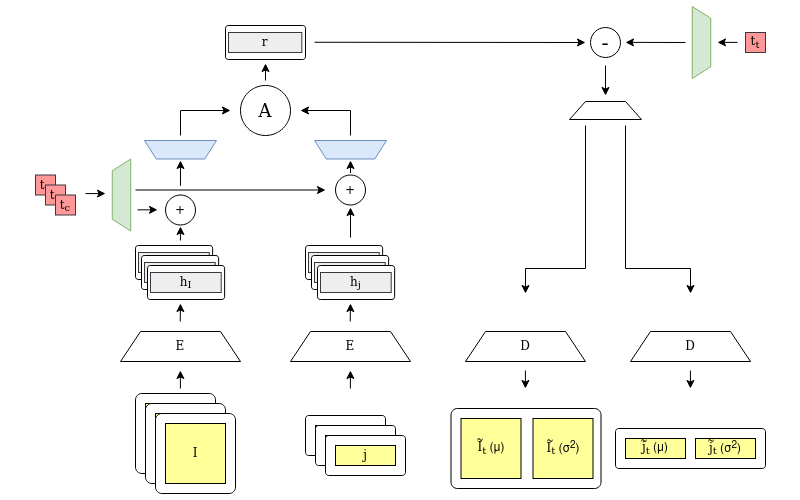}
          \subcaption{Proposed DMBN-CI architecture}
          \label{fig:dmbn-ci-architecture}
      \end{subfigure}
  
  \caption{DMBN architecture, adapted from \cite{seker2022imitation}. The same color convention as in \cite{garnelo2018conditional} has been adopted to indicate inputs (yellow) and outputs (red). Networks with the same color share weights.}
  \label{fig:architecture}
\end{figure*}

The Deep Modality Blending Network \cite{seker2022imitation} is a multimodal network designed to reconstruct a complete action sequence given a partial observation. In the original setup, the network takes input from a 7-degree-of-freedom robot arm, collecting images from a fixed viewpoint and its joint values, Fig. \ref{fig:dmbn-architecture}$.i$. The data consists of sequences of grasping and pushing motions performed by the robot arm in a simulation environment.

The network encodes the two modalities separately using convolutional and fully connected neural networks, for the image and proprioceptive input respectively, Fig. \ref{fig:dmbn-architecture}$.ii$. The encoded features are then combined through averaging, Fig. \ref{fig:dmbn-architecture}$.iii$, generating a shared multimodal hidden representation. This is then augmented with target times, Fig. \ref{fig:dmbn-architecture}$.iv$, and is passed through a separate decoder for each modality, Fig. \ref{fig:dmbn-architecture}$.v$, generating both mean and variance signals, Fig. \ref{fig:dmbn-architecture}$.vi$, reconstructing a distribution over the target set.
%The DMBN is trained by backpropagation \cite{rumelhart1986learning}, minimizing the Gaussian Negative Log Likelihood Loss \cite{nix1994estimating} between the generated distribution and the desired target.

% add figure similar to that in the original paper

% Deep dive into how it works

% - describe the architecture in detail
% - detour on what CNP (and NPs in general) are
% - stress that the allegedly Mirror-Neuron like property won't be addressed here
%   (should mention that it might spur from underfitting observations?)
% - multimodal only in  the NPF space, the two signals are otherwise encoded and decoded independently. More study on biological data needed here

\section{Model Evaluation} \label{sec:evaluation}
% mention that you've reimplemented the net, with small adjustements necessary to replicate, partially, the original results

To conduct this study the DMBN has been reimplemented\footnote{Source code: https://gitlab.iit.it/cognitiveInteraction/dmbn-torch} from scratch in PyTorch \cite{paszke2017automatic}, with some necessary adjustments in order to replicate the original results\footnote{The major one being splitting each decoder head in two, separating the mean and variance output}.
%A first evaluation, described in more details below, demonstrated the inability of the net to accurately model time relations in the data, which evolved in the adoption of a different process through which the time information is encoded and appended to the input and the set-representing element.

\subsection{Time as Channel}

\begin{figure}[t]
  \centering
  \begin{minipage}[b]{0.3\linewidth}
    \centering
    \includegraphics[width=\linewidth]{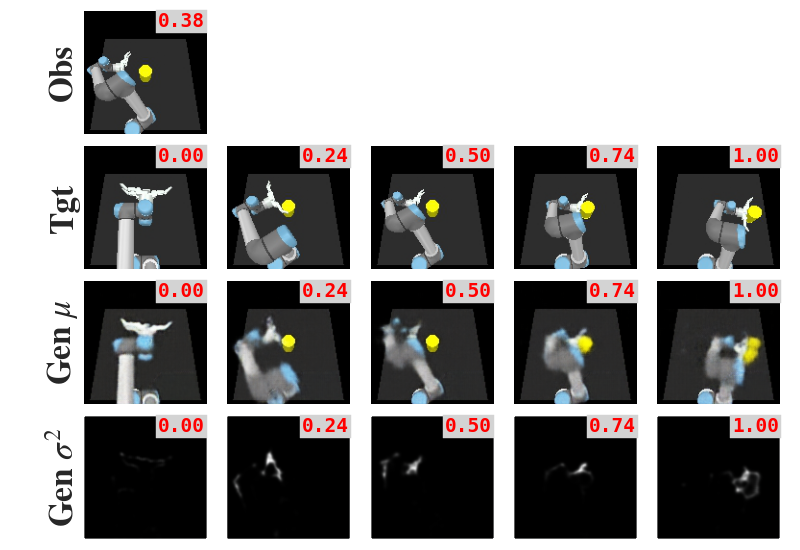}
  \end{minipage}
  \hfill
  \begin{minipage}[b]{0.3\linewidth}
    \centering
    \includegraphics[width=\linewidth]{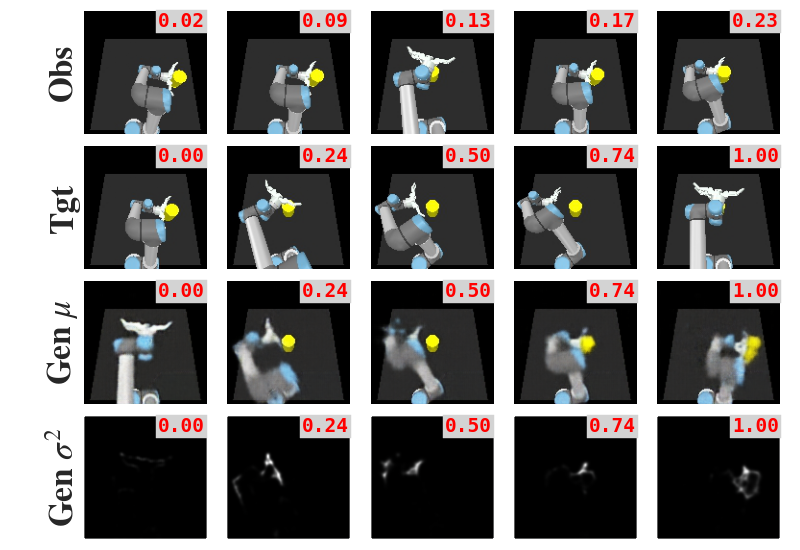}
  \end{minipage}
  \hfill
  \begin{minipage}[b]{0.3\linewidth}
    \centering
    \includegraphics[width=\linewidth]{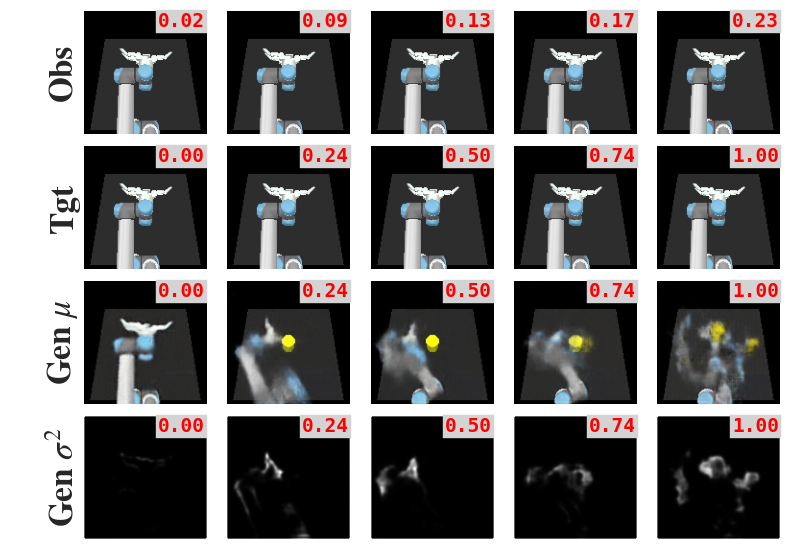}
  \end{minipage}
  
  \begin{minipage}[b]{0.3\linewidth}
    \centering
    \includegraphics[width=\linewidth]{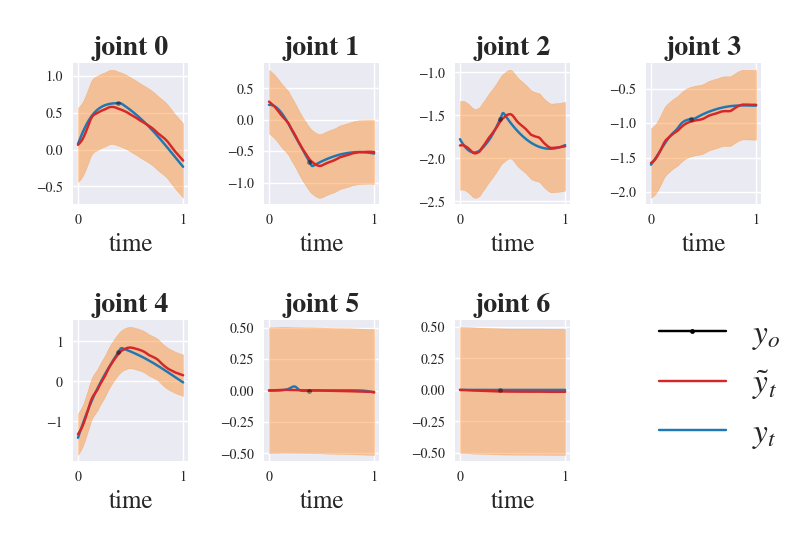}
    \subcaption{1 observation, t-sequence}
    \label{fig:dmbn-out:t}
  \end{minipage}
  \hfill
  \begin{minipage}[b]{0.3\linewidth}
    \centering
    \includegraphics[width=\linewidth]{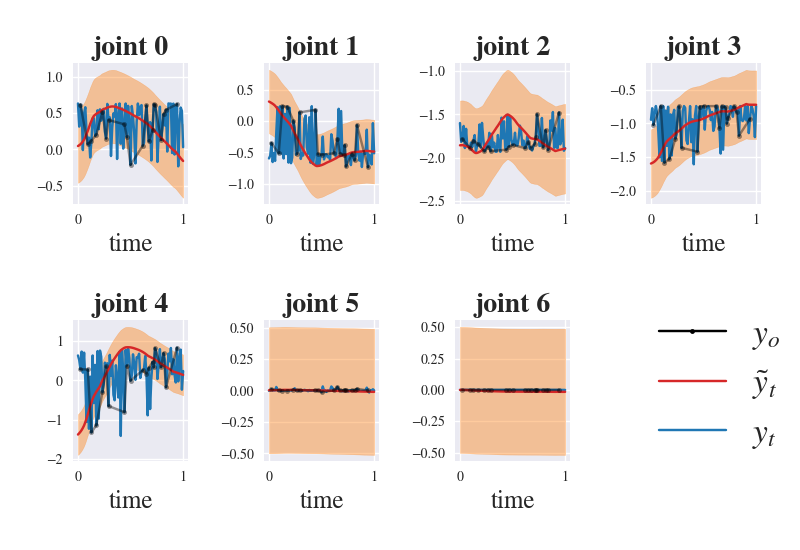}
    \subcaption{20 observations, p-sequence}
    \label{fig:dmbn-out:p}
  \end{minipage}
  \hfill
  \begin{minipage}[b]{0.3\linewidth}
    \centering
    \includegraphics[width=\linewidth]{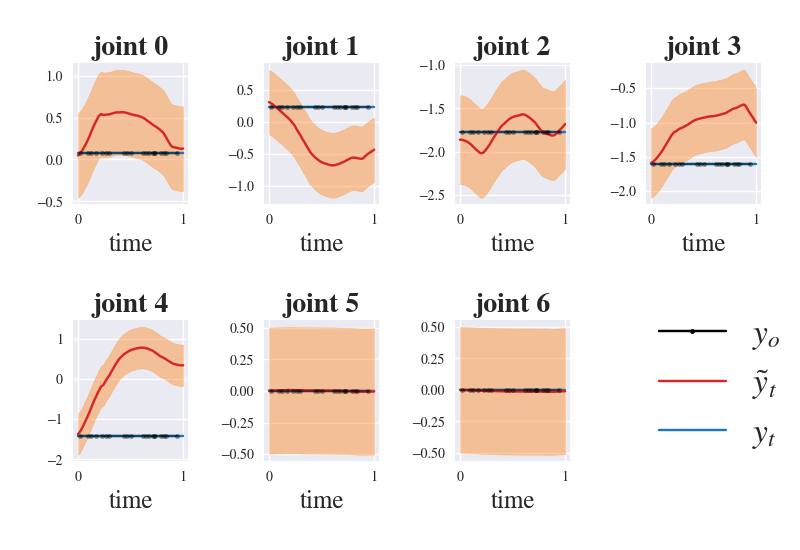}
    \subcaption{20 observations, f-sequence}
    \label{fig:dmbn-out:f}
  \end{minipage}
  
  \caption{Generated bimodal output by the original DMBN architecture.\\Legend: $y_o$, observation; $\tilde y_t$, generated; $y_t$, target}
  \label{fig:dmbn-out}
\end{figure}

In the original formulation, the context time is inserted as a new channel in both the visual and proprioceptive inputs, Fig. \ref{fig:dmbn-architecture}$.i$, while target times are appended to the representative element of the context set, Fig. \ref{fig:dmbn-architecture}$.iv$. 
Fig. \ref{fig:dmbn-out:t} shows the output of the original DMBN on a test sequence (\textit{t-sequence}) demonstrating a qualitative good reconstruction despite a general overestimation of the variance. 
%A couple of remarks can be made on the generated variance. In the visual modality, a higher value is focused on the end-effector of the robotic arm: while this naturally emerges from the higher dynamic nature of this part, it is interesting to notice the similarity with the attentive bias for own hands in infants \cite{white1964observations}. In the joint time-series it is instead clear that the net fails at representing the last two joints, becoming overconfident in the generated variance, possibly due to their small dynamic range.
To evaluate the model's capabilities on diverse data, synthetic sequences were generated by permuting\footnote{The permutation refers actually to the time instants associated to each observation, not on the order of the context elements themselves, that is of course ignored by design by the set operator.} and freezing the test sequence (\textit{p-sequence} and \textit{f-sequence} respectively).
Due to the mismatch between semantic and time content, we would expect a nonsensical output to the observed p-sequence.
From Fig. \ref{fig:dmbn-out:p} we observe that the net, despite being presented with a motion that jumps forward and backward in time, reconstructs the "ordered" underlying sequence. It behaves as if the net were ignoring the time of each observation, looking only at its content.
To quantitatively verify this claim, FCN heads were trained on frozen DMBN encoders, to regress back to the context time sequence, effectively inverting the encoding process for it. Table \ref{tab:time-regression} compares the network's performance with two baseline models: untrained encoders (\textit{Random}) and encoders receiving a null time signal (\textit{Null}). The DMBN performs worse than the random baselines, confirming that the temporal information is actively silenced by the net.
% here describe the total lack of time information in the original network
% - start by trying to scramble time -> notice no effect on reconstruction
% - table with results of time regression compared with random net and time = 0 always

\subsection{Time as Context}

\begin{figure}[t]
  \centering
  \begin{minipage}[b]{0.3\linewidth}
    \centering
    \includegraphics[width=\linewidth]{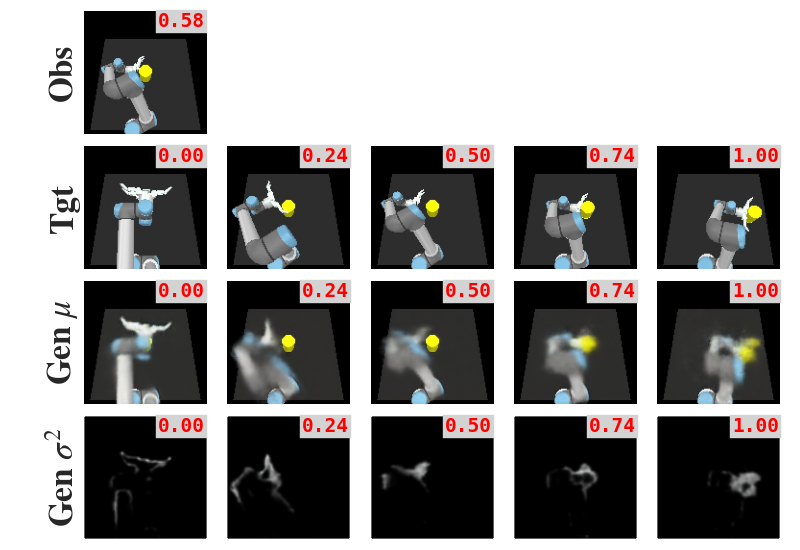}
  \end{minipage}
  \hfill
  \begin{minipage}[b]{0.3\linewidth}
    \centering
    \includegraphics[width=\linewidth]{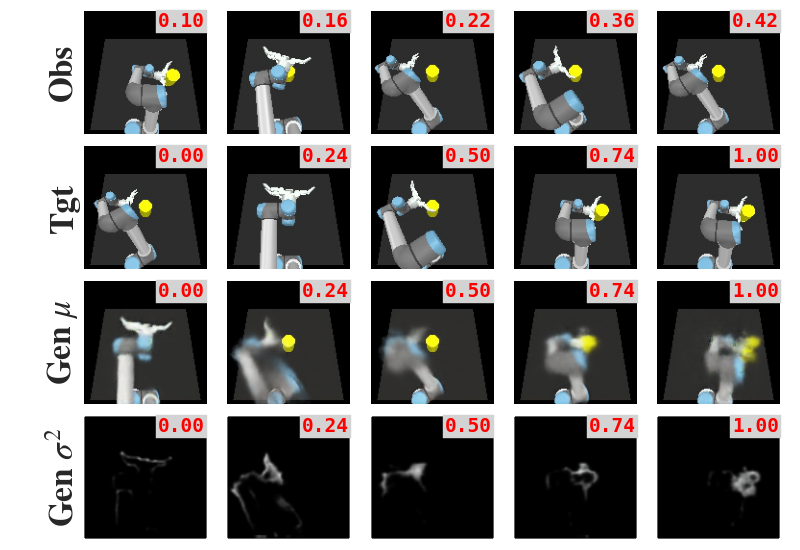}
  \end{minipage}
  \hfill
  \begin{minipage}[b]{0.3\linewidth}
    \centering
    \includegraphics[width=\linewidth]{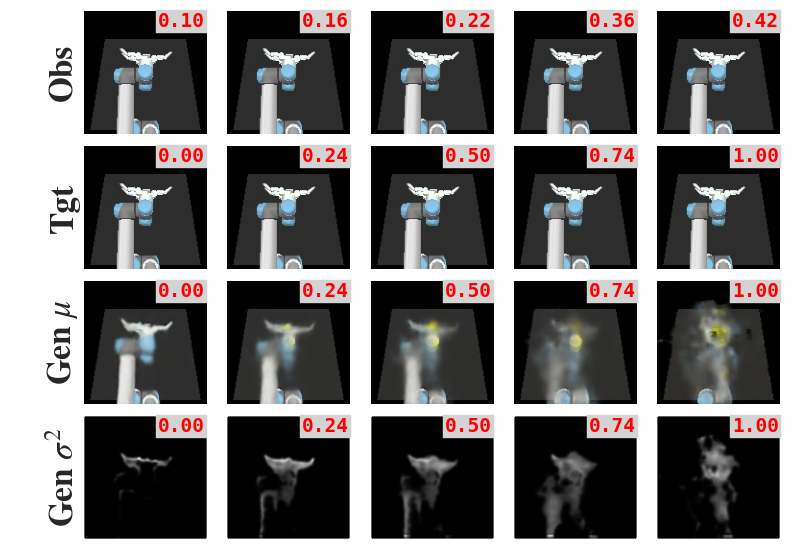}
  \end{minipage}
  
  \begin{minipage}[b]{0.3\linewidth}
    \centering
    \includegraphics[width=\linewidth]{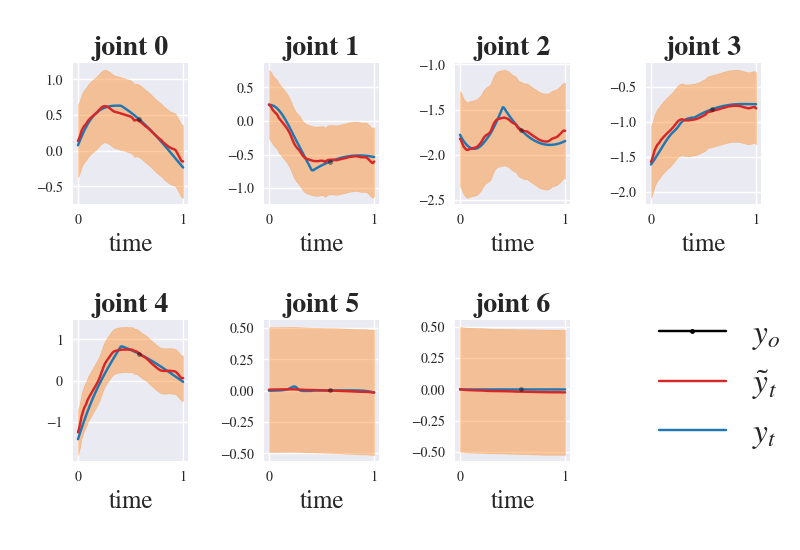}
    \subcaption{1 observation, t-sequence}
    \label{fig:dmbn-ci-out:t}
  \end{minipage}
  \hfill
  \begin{minipage}[b]{0.3\linewidth}
    \centering
    \includegraphics[width=\linewidth]{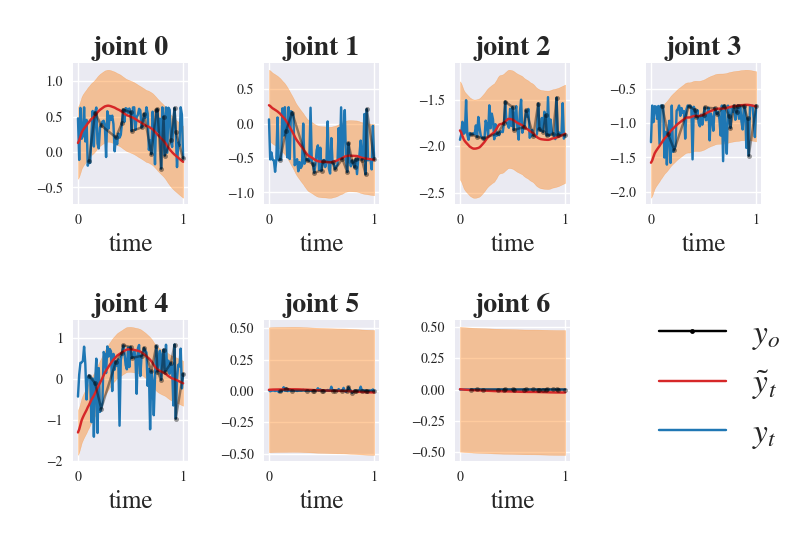}
    \subcaption{20 observations, p-sequence}
    \label{fig:dmbn-ci-out:p}
  \end{minipage}
  \hfill
  \begin{minipage}[b]{0.3\linewidth}
    \centering
    \includegraphics[width=\linewidth]{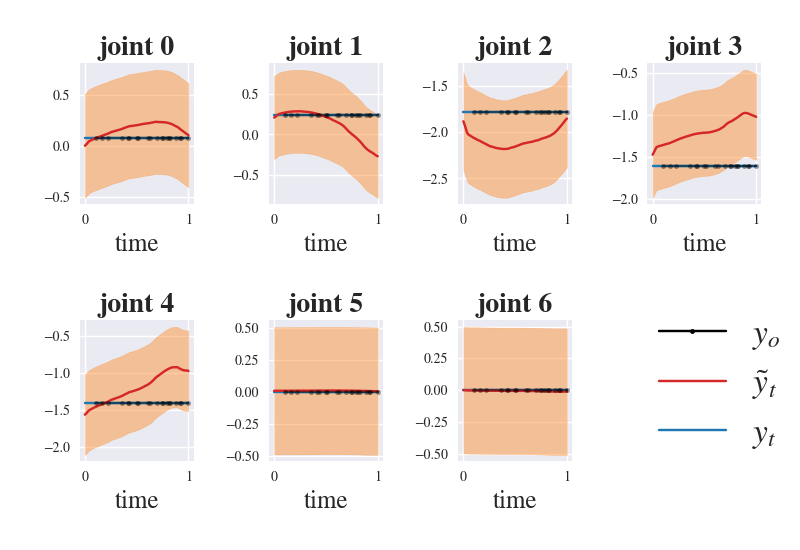}
    \subcaption{20 observations, f-sequence}
    \label{fig:dmbn-ci-out:f}
  \end{minipage}
  
  \caption{Generated bimodal output by the proposed DMBN-PTE architecture}
  \label{fig:dmbn-ci-out}
\end{figure}

Temporal information is however necessary for the CNP to predict the dynamics of the observed motion.
To reintroduce time in the hidden layer, inspiration was taken from positional encodings in Transformer networks \cite{vaswani2017attention}. In the modified architecture, the temporal information is projected with a 1-layer Fully Connected Network (FCN) onto the hidden space, \textit{added} to the encodings, and the result is non-linearly projected on the same space. Target times, after going through the same transformation, are instead \textit{subtracted} from the set-representing elements before decoding\footnote{The rationale being to revert the process of going from a $signal$ to a $signal+time$ space in the time insertion.}. 
Data in Table \ref{tab:time-regression} supports the presence of time in the hidden layer, given the order of magnitude smaller losses compared to the \textit{Random} net.
This updated architecture, referred to as DMBN-Positional Time Encoding (DMBN-PTE), is depicted in Fig. \ref{fig:dmbn-ci-architecture}. The training dataset is augmented by randomly modifying the speed of sections of the sequence and adding repeated frames.
Fig. \ref{fig:dmbn-ci-out} shows the generated output for the \textit{t-}, \textit{p-}, and \textit{f-sequences}. The DMBN-PTE manages to capture the freezing in the f-sequence for almost the entire duration, but once again reconstructs an unexpectedly ordered signal for the p-sequence, which we hypothesize is a sign of underfitting of the context set or excessive memorization.

\begin{table*}
  \scriptsize
  \caption{Regression to context time performances. The results for both modality encoders are reported with 95\% confidence interval in parentheses.}
  \label{tab:time-regression}
  \begin{tabular}{crr}
    \toprule
    Model & Image Encoder Loss (1e-3) & Joint Encoder Loss (1e-3)\\
    \midrule
    Null & 42.95 (29.05, 56.85) & 88.52 (81.76, 95.29)\\
    Random & 2.18 (1.80, 2.56) & 4.48 (3.15, 5.81)\\
    DMBN & 29.63 (25.47, 33.79) & 86.56 (79.18, 93.94)\\
    DMBN-PTE & 0.69 (0.64, 0.74) & 0.19 (0.15, 0.23)\\
  \bottomrule
\end{tabular}
\end{table*}

% describe the insertion of time as additive information on the hidden space
% (remember to mention reprojection and normalization, otherwise the averaging would loose ordering information
%  - point to results in that same table as before
% - how does scramble time behave?

%\subsection{Incremental Sequence Length}
% although non autoregressive, learning can be facilitated by slowly enlarging the subset of observation and target points

% \subsection{Random Time Shift}
% to lower the association of a frame with its position in time, with the aim to help the system generalize, the subsequences were randomly shifted in time
\section{Future Work}

Further investigations are needed to evaluate the suitability of CNPs for representing high-dimensional and multimodal time series. The potential directions for future work include:

\begin{itemize}
    \item Richer Datasets: exploring ecological datasets, such as the BAIR Robot Pushing \cite{ebert2017self}, could lower the need for synthetic data augmentation and provide stronger time representations.
    \item Different Neural Processes Architectures: the structural biases of alternative NP networks could simplify learning meaningful features, such as the relative encoding of the input signal by ConvolutionalCNPs \cite{gordon2019convolutional}.
    \item Latent Path: The literature on Video Prediction incorporates random latent variables in deterministic pathways to capture non-deterministic information, not dissimilar from the proposal in the original NP work \cite{garnelo2018neural}.
    \item Online Learning: learning while interacting with a simulated or real environment could lead to stronger multimodal consistency. The generated uncertainty could drive exploratory behaviors, facilitating faster coverage of the state space, a concept known as "Curiosity-Driven Exploration" in the Reinforcement Learning literature \cite{blau2019bayesian}.
\end{itemize}
\section{Conclusion}

The non-autoregressive and probabilistic nature of DMBNs poses them as good candidates for a MNS-inspired robotic architectures. However, this preliminary evaluation revealed that their original formulation struggles to represent time in a way that facilitates the generalization of the dynamics of the learned visual-motor correlation. The proposed modifications to the network and training procedure have shown promising results, but are just the first step in successfully applying Neural Processes to multimodal, high-dimensional, time-series predictions.
%\section{Appendices}

\begin{acknowledgments}
     We gratefully acknowledge the HPC infrastructure and the Support Team at Fondazione Istituto Italiano di Tecnologia. This research has been conducted in the framework of a Starting Grant from the European Research Council (ERC) under the European Union’s Horizon 2020 research and innovation programme. G.A. No 804388, wHiSPER.
\end{acknowledgments}

%%
%% Define the bibliography file to be used
\bibliography{sample-ceur}

%\appendix
%\input{Sections/appendix_net_implementation}

\end{document}